\documentclass[runningheads]{llncs}
\usepackage{graphicx}
\usepackage{subcaption}
\usepackage{xcolor}
\usepackage{algorithm}
\usepackage{algpseudocode}
\usepackage{graphicx}
\usepackage{textcomp}
\usepackage{amsmath}
\usepackage{xcolor}
\usepackage{booktabs}
\usepackage[utf8]{inputenc}
\usepackage{float}
\usepackage{amsfonts}
\usepackage{subcaption}
\begin{document}
\title{MT3DNet: Multi-Task learning Network for 3D Surgical Scene Reconstruction}
%
%
\author{Mithun Parab$^*$\inst{1} \and
Pranay Lendave$^*$\inst{2} \and
Jiyoung Kim \inst{3} \and
Thi Quynh Dan Nguyen \inst{4} \and 
Palash Ingle\inst{3}
}
%
%
\institute{R.J. College,  Mumbai, India \and
K J Somaiya College of Engineering, Mumbai, India \and
Sejong University, Seoul, South Korea  \and
Korean institute of Oriental Medicine, Daejeon , South Korea
}


%
\maketitle              
\begin{abstract}
In image-assisted minimally invasive surgeries (MIS), understanding surgical scenes is vital for real-time feedback to surgeons, skill evaluation, and improving outcomes through collaborative human-robot procedures. Within this context, the challenge lies in accurately detecting, segmenting, and estimating the depth of surgical scenes depicted in high-resolution images, while simultaneously reconstructing the scene in 3D and providing segmentation of surgical instruments along with detection labels for each instrument. To address this challenge, a novel Multi-Task Learning (MTL) network is proposed for performing these tasks concurrently. A key aspect of this approach involves overcoming the optimization hurdles associated with handling multiple tasks concurrently by integrating a Adversarial Weight Update into the MTL framework, the proposed MTL model achieves 3D reconstruction through the integration of segmentation, depth estimation, and object detection, thereby enhancing the understanding of surgical scenes, which marks a significant advancement compared to existing studies that lack 3D capabilities. Comprehensive experiments on the EndoVis2018 benchmark dataset underscore the adeptness of the model in efficiently addressing all three tasks, demonstrating the efficacy of the proposed techniques.

\end{abstract}
\def\thefootnote{$*$}\footnotetext{These authors contributed equally to this work}\def\thefootnote{\arabic{footnote}}
\section{Introduction}
\label{sec:intro}
Efficient 3D reconstruction for surgical scene understanding of MIS environments is crucial during robotic surgery since it enhances precision, reliability and reduces trauma and recovery time. However, surgeons can still face workflow disruptions due to limited tactile feedback or system issues. Moreover, the complexity of surgical environments, characterized by factors such as smoke, bodily fluids, varying lighting conditions, and partial occlusions, poses significant challenges to image interpretation \cite{pakhomov2019deep}. Hence, there arises a pressing need for 3D reconstruction of surgical scenes, it enhances surgical performance, offers real-time feedback, aids in skill assessment for novice surgeons, and enables detailed analysis of tool movements. Recognizing the limitations of single-task learning (STL) for individual tasks like semantic segmentation \cite{wang2021efficient}, instance segmentation \cite{kurmann2021mask}, and object detection \cite{cheng2021deep} in surgical scene understanding, a multi-task learning (MTL) approach is preferred that performs all three tasks concurrently, promoting computational efficiency.
However, recent breakthroughs in MTL haven't addressed depth estimation, which is crucial for 3D reconstruction, leaving a gap in achieving truly comprehensive scene understanding. There are approaches for 3D reconstruction by finding disparity maps using stereo camera setup \cite{psychogyios2022msdesis}, but these approaches adds an additional camera which is not feasible in hardware and space constrained MIS procedures. Thus, this approach emphasizes on monocular depth estimation using a single camera view that can be used for 3D reconstruction of the MIS environment.

The progress in comprehending surgical scenes heavily relies on the EndoVis2018 datasets, which feature annotated robotic instrument data for tasks like segmentation and detection. Despite containing seven tube-like instruments, these datasets lack monocular depth annotations for MIS settings. To address this, the authors employed the state-of-the-art STL depth estimation model, Depth Anything\cite{depth_anything_v1}, to generate depth maps for training the MTL model.
\\{Our Contributions:}
\begin{itemize}
    \item Develop a novel multi-task framework for performing three tasks: segmentation, instrument detection, and monocular depth estimation concurrently.
    \item Implementation of a proficient 3D reconstruction process for surgical scenes and instruments, using depth estimation from the Multi-Task Learning model.
    \item Incorporation of a Adversarial Weight Update within our methodology, strategically employed to achieve Pareto optimality for each designated task.
\end{itemize}

\section{Related Work}
\label{sec:related-work}
The use of object detection, segmentation, tracking, and depth estimation in medical imaging spans various surgical and diagnostic specialties. Diverse methodologies leverage datasets like the EndoVis, with approaches including single-task learning (STL). Furthermore, 3D reconstruction \cite{psychogyios2022msdesis} plays a crucial role in surgical understanding. Our focus lies on multitask learning for surgical instrument tasks, recognizing its benefits in generalization and computational efficiency.

\subsection{Segmentation and object detection}
\label{subsec:seg}
CNN models are utilized for binary, semantic, and instance segmentation in surgical tool tracking. TernausNet \cite{iglovikov2021ternausnet}, excels in binary segmentation, while ISINet \cite{zhao2022trasetr}, TraSeTR \cite{zhao2022trasetr}, and AP-MTL \cite{islam2020ap} focus on instance segmentation. Considering the computational demands, lighter techniques like binary segmentation can enhance processing speed. Moreover, combining binary segmentation with object detection can yield comparable outcomes to instance and semantic segmentation methods. Advancements in instrument detection include TernausNet-16 \cite{iglovikov2021ternausnet}, a CNN extension of U-Net, and methods by Kurmann et al. \cite{kurmann2017simultaneous} for 2D vision-based recognition and Sarikaya et al. \cite{sarikaya2017detection} for region proposal network (RPN) and multi-modal two-stream convolutional network. S3Net \cite{baby2023forks} employs instance segmentation for detection, but integrating binary segmentation with detection readings offers a more cost-effective approach to achieve instance segmentation.

\subsection{Depth Estimation}
\label{subsec:depth-estimation}
Traditionally hampered by data scarcity, stereo-based 3D reconstruction relied on handcrafted methods \cite{stoyanov2010real}, \cite{zhou2019real} validated on specialized datasets \cite{eddie2020serv}. However, new advancements in monocular depth estimation now rival the capabilities of stereo \cite{psychogyios2022msdesis} and LiDAR setups. This ensures cost-effectiveness and system flexibility, while harnessing the latest developments to provide precise visual guidance in robotic surgery. MTL leverages semantic segmentation for scene context in monocular depth estimation, addressing issues like blurry object boundaries and enhancing accuracy, as pioneered in \cite{eigen2015predicting}. MTL networks optimize efficiency by sharing a single feature extraction backbone, cutting down on inference time and memory usage, enabling models like SENSE \cite{jiang2019sense} to predict multiple outputs concurrently.

\subsection{Multi-Task Learning}
\label{subsec:mtl}
In computer vision, various MTL models like MaskRCNN \cite{he2017mask} have been devised for concurrent semantic segmentation and object detection tasks. Recently, efforts have been made by UberNet \cite{kokkinos2017ubernet} and AP-MTL \cite{islam2020ap} to refine MTL models through multi-phase training approaches and fine-tuning. However, while MTL has seen significant progress in segmentation and detection tasks, it trails behind in depth estimation for comprehensive 3D reconstruction. MSDESIS \cite{psychogyios2022msdesis} attempted multitask depth estimation and segmentation utilizing a stereo camera setup, which could present challenges in hardware and space-limited MIS procedures. Therefore, additional research is required to explore monocular multi-task learning approaches in robotic MIS environments.

\section{Methodology}
\label{sec:methdology}

\begin{figure}[t]
    \begin{center}
    \includegraphics[width=1\textwidth]{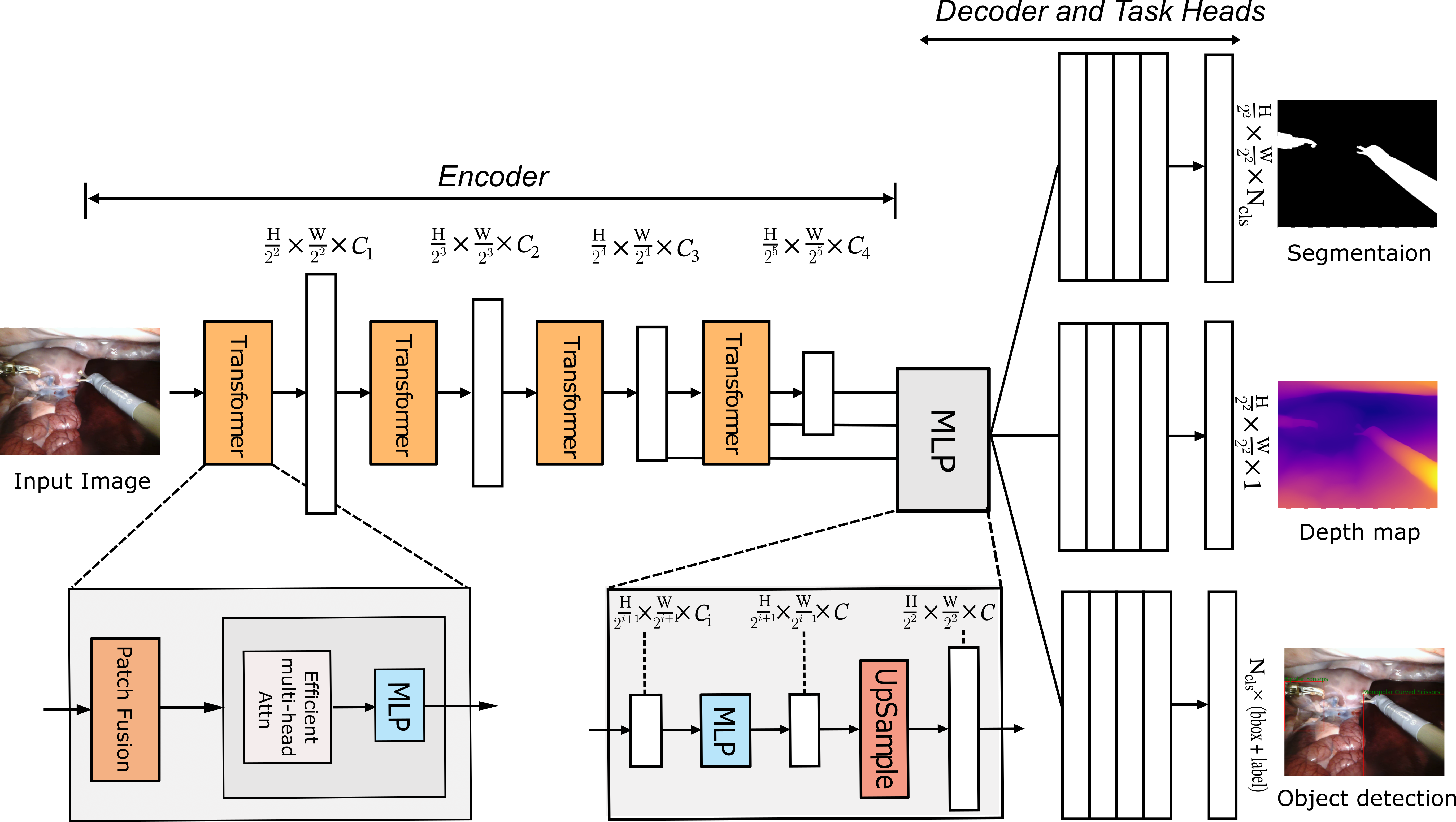}
    \caption{The proposed Multi-Task Learning Transformer (MT3DNet) architecture.}
    \label{fig:mt3dnet-arch}
    \end{center}
\end{figure}

The proposed model architecture comprises an Encoder, Decoder, and task-specific heads, integrated into a unified Multi-Task Learning Transformer (MT3DNet) framework is depicted in Fig. \ref{fig:mt3dnet-arch}.

\subsubsection{Encoder.}
\label{subsubsec:encoder}
The encoder architecture is designed to extract hierarchical features from input data by progressively transforming and downsampling it through multiple stages. Each stage in the encoder consists of a series of operations that refine the representation, leveraging both convolutional and attention-based mechanisms. The stages are stacked sequentially, each having a defined number of layers and different configurations for patch size, overlap, and dimensionality reduction. At the core of the encoder, each stage performs patch merging, where the spatial resolution of the input is reduced while preserving essential features. This is achieved using the concept of overlap patch merging, where overlapping patches are combined to enhance the information retention during downsampling. The input is then passed through a series of encoder blocks, each block consisting of self-attention and feed-forward layers. These blocks are meant to capture long-range dependencies and intricate patterns in the data.

The depth of the encoder determines the number of encoder blocks per stage, allowing the network to progressively refine its features at different levels of abstraction. Each encoder block is then equipped with a regularization technique, controlled by a dynamic drop path probability, which aids the learning process by randomly dropping certain paths during the training process. This helps prevent overfitting and promotes better generalization. The output of each stage of the encoder is passed through a normalization layer to stabilize learning and ensure consistent feature scaling across the network. The multi-stage design allows the encoder to capture both local and global information with each stage progressively handling larger contexts and more abstract features.

\subsubsection{Decoder.}
\label{subsubsec:decoder}
The Decoder component of the architecture takes as input a set of features ($F$) with dimensions corresponding to the batch size, number of channels, height, and width. It processes these features and outputs a transformed set of features ($F'$) with the same spatial and channel dimensions. The spatial size of the output features is fixed to a quarter of the initial spatial size of the first features. This Decoder component typically follows the Encoder in the architecture and receives the output from the Encoder as its input. It plays a crucial role in further processing the encoded features and preparing them for use in downstream tasks.

\subsubsection{Task Heads.}
\label{subsubsec:taskheads}
After passing through the Decoder, the processed features are fed into task-specific heads, which are responsible for reducing the feature dimensionality to match the requirements of individual tasks. These task heads are tailored according to the specific task at hand, such as segmentation, monocular depth estimation, or object detection. Each task head receives its input tensor from the Decoder and performs the final processing before producing the output for the respective task. This task head stage represents the final block of the Multi-Task Learning (MTL) architecture.

\subsubsection{Multi-Task Learning Transformer (MT3DNet).}
\label{subsubsec:mtl-mt3dnet}
MT3DNet is a comprehensive architecture that integrates Encoder, Decoder, and task heads into a unified framework for multi-task learning. Leveraging Transformer-based architectures, MT3DNet efficiently processes input data and addresses multiple tasks simultaneously. The Encoder utilizes Transformer blocks to effectively encode input data, capturing intricate relationships. The subsequent Decoder refines encoded features for task-specific processing. Task heads provide modular and task-specific feature representations, optimizing performance for each task. This integration offers a versatile and efficient solution for multi-task learning across domains and applications.




\section{Loss and Evaluation Metrics}
\label{sec:loss-eval}
\subsection{Losses}
\label{subsec:loss}

In the Multi-Task Learning (MTL) paradigm, the total loss function \eqref{eqn:total-loss} encompasses individual losses from segmentation \eqref{eqn:seg}, depth estimation \eqref{eqn:depth}, and object detection \eqref{eqn:objdet} tasks.
\begin{equation}
\label{eqn:total-loss}
\text{Total Loss} \left ( \mathcal{L}_{total}  \right) = w_{1} \times \mathcal{L}_{seg} +  w_{2} \times \mathcal{L}_{depth} +  w_{3} \times \mathcal{L}_{detection}
\end{equation}
where, $w_1$, $w_2$ and $w_3$ are weights for each task's loss.
\begin{itemize}
       \item \textbf{Segmentation Loss:} The segmentation loss combines binary cross-entropy (BCE) or cross-entropy (CE) loss with the mean Intersection over Union (mIoU) loss to handle both binary and multi-class segmentation tasks. The general form is given as:
    
    \begin{equation}
    \label{eqn:seg}
    \mathcal{L}_{seg} = \alpha \cdot \mathcal{L}_{CE/BCE} + \beta \cdot (1 - \text{mIoU}),
    \end{equation}
    
    where:
    \begin{itemize}
        \item $\mathcal{L}_{CE/BCE}$ represents the cross-entropy loss for multi-class or binary cross-entropy loss for binary segmentation.
        \item $\text{mIoU}$ is the mean Intersection over Union, defined as:
        \begin{equation}
        \text{mIoU} = \frac{\sum_{c} \frac{\text{Intersection}_c}{\text{Union}_c}}{C},
        \end{equation}
        where $C$ is the number of classes, $\text{Intersection}_c$ is the overlapping area between predicted and ground truth for class $c$, and $\text{Union}_c$ is their union.
        \item $\alpha$ and $\beta$ are weighting factors for balancing the loss components.
    \end{itemize}
    
    For binary segmentation, the BCE loss is computed as:
    \begin{equation}
    \mathcal{L}_{BCE} = -\frac{1}{N} \sum_{i=1}^{N} \left[ y_i \log(p_i) + (1 - y_i) \log(1 - p_i) \right],
    \end{equation}
    where $y_i$ is the ground truth, $p_i$ is the predicted probability, and $N$ is the number of pixels.
    
    For multi-class segmentation, the CE loss is:
    \begin{equation}
    \mathcal{L}_{CE} = -\frac{1}{N} \sum_{i=1}^{N} \sum_{c=1}^{C} y_{i,c} \log(p_{i,c}),
    \end{equation}
    where $y_{i,c}$ and $p_{i,c}$ are the ground truth and predicted probability for class $c$ at pixel $i$.
    
    The combined loss ensures both accurate pixel-wise predictions and effective region-wise overlap.

   \item \textbf{Depth Loss:}
    The depth loss is defined as a weighted combination of three components: SSIM loss, edge loss, and mean absolute error (MAE). Let \( D_p \) and \( D_t \) represent the predicted depth and target depth, respectively. The total depth loss \( \mathcal{L}_{\text{depth}} \) is given by:
    
    \begin{equation}
    \label{eqn:depth}
    \mathcal{L}_{\text{depth}} = w_1 \cdot \mathcal{L}_{\text{SSIM}} + w_2 \cdot \mathcal{L}_{\text{edge}} + w_3 \cdot \mathcal{L}_{\text{MAE}},
    \end{equation}
    
    where \( w_1, w_2, \) and \( w_3 \) are the weights assigned to each loss component. The components are defined as follows:
    
    \begin{itemize}
        \item \textbf{SSIM Loss:}
        \begin{equation}
        \mathcal{L}_{\text{SSIM}} = 1 - \text{SSIM}(D_p, D_t),
        \end{equation}
        where \( \text{SSIM}(\cdot, \cdot) \) is the structural similarity index function.
    
        \item \textbf{Edge Loss:}
        The edge loss is computed using a Sobel filter \( \mathbf{S} \) to extract gradients from the predicted and target depth maps:
        \begin{equation}
        \mathcal{L}_{\text{edge}} = \frac{1}{N} \sum_{i=1}^{N} \left| \mathbf{S}(D_p)_i - \mathbf{S}(D_t)_i \right|,
        \end{equation}
        where \( N \) is the number of pixels in the depth map.
    
        \item \textbf{Mean Absolute Error (MAE):}
        \begin{equation}
        \mathcal{L}_{\text{MAE}} = \frac{1}{N} \sum_{i=1}^{N} \left| D_p^i - D_t^i \right|.
        \end{equation}
    \end{itemize}

    \item \textbf{Object Detection Loss:} The object detection module integrates  \text{Smooth $\mathcal{L}_{1}$}  Loss for bounding box localization and cross-entropy for classification.
   \begin{equation}
   \label{eqn:objdet}
    \text{Object Detection Loss} \left ( \mathcal{L}_{detection} \right ) = \text{Smooth $\mathcal{L}_{1}$ Loss} + \text{CE}
   \end{equation}
\end{itemize}

\subsection{Adversarial Weight Updates for Multi-Task Learning}

In multi-task learning, tasks often exhibit conflicts, making balanced optimization crucial. We propose an adversarial approach that dynamically adjusts task weights to achieve equilibrium, inspired by Multiplicative Weights Update (MWU). This ensures tasks contribute proportionally to their significance, minimizing conflicts while improving overall performance. Below, we present the core algorithm for this process.

To further illustrate this, the adversarial updates of the weights make use of loss values across tasks, driving optimization. It adaptively amplifies the higher loss tasks so that low performers get more attention in the training process. Details for this are available in Algorithm~\ref{alg:adversarial_weight_update}.

\begin{algorithm}[H]
\caption{Adversarial Weight Update for Multi-Task Learning}
\begin{algorithmic}[1]
\Require Task losses $\{\ell_t\}_{t=1}^T$, initial weights $\{w_t\}_{t=1}^T$, learning rate $\eta$, small constant $\epsilon$
\Ensure Updated weights $\{w_t^{\text{new}}\}_{t=1}^T$
\State Compute updated weights:
\[
w_t^{\text{new}} \gets w_t \cdot \exp(-\eta \cdot \ell_t) \quad \forall t \in \{1, \ldots, T\}
\]
\State Normalize weights:
\[
w_t^{\text{new}} \gets \frac{w_t^{\text{new}}}{\sum_{k=1}^T w_k^{\text{new}} + \epsilon} \quad \forall t \in \{1, \ldots, T\}
\]
\end{algorithmic}
\label{alg:adversarial_weight_update}
\end{algorithm}

The optimal weight computation method accompanies the adversarial update method. This method lines up task gradients to diminish inter-task conflicts and stabilize learning dynamics. The gradient alignment method makes sure that the learning dynamics are robust against any kind of change in the task complexity. Algorithm~\ref{alg:optimal_weight_computation} describes the procedure.

\begin{algorithm}[H]
\caption{Optimal Weight Computation via Gradient Alignment}
\begin{algorithmic}[1]
\Require Task gradients $\mathbf{G} \in \mathbb{R}^{T \times P}$, regularization term $\lambda$, unit vector $\mathbf{1} \in \mathbb{R}^T$
\Ensure Optimal weights $\mathbf{w}^*$
\State Compute gradient Gram matrix:
\[
\mathbf{G}^\top \mathbf{G} \gets \mathbf{G}^\top \mathbf{G} + \lambda \mathbf{I}
\]
\State Solve for optimal weights:
\[
\mathbf{w}^* \gets \arg\min_{\mathbf{w}} \| \mathbf{G}^\top \mathbf{w} - \mathbf{1} \|^2
\]
\State Normalize weights:
\[
w_t^* \gets \frac{w_t^*}{\sum_{k=1}^T w_k^* + \epsilon} \quad \forall t \in \{1, \ldots, T\}
\]
\end{algorithmic}
\label{alg:optimal_weight_computation}
\end{algorithm}

The use of these weight updates drives this multi-task learning framework to reliably align gradients and optimize the tasks without explicit optimization dynamics, hence being robust and efficient in the overall learning process.

\begin{table}[H]
\centering
\caption{Performance of different models for segmentation (Seg.) and object detection (Det.) tasks, with metrics including Dice, and mAP.}
\label{tab:seg-obj}
\begin{tabular}{@{}ccccc@{}}
\toprule
Model                                      & Regime       & Arch.      & Segmentation & Object detection \\ \midrule
                                           &              &            & Dice         & mAP              \\ \midrule \midrule
TernausNet11\cite{iglovikov2021ternausnet} & Seg.         &            & 0.920        &                  \\
ERFNet\cite{romera2017erfnet}              & Seg.         & ERFNet     & 0.923        & -                \\
LinkNet\cite{chaurasia2017linknet}         & Seg.         & UNet       & 0.941        & -                \\
DeepLabv3+\cite{baheti2020semantic}        & Seg.         & ResNet     & 0.947        & -                \\
Yolov3\cite{redmon2018yolov3}              & Det.         & Darknet-53 & -            & 0.363            \\
SSD\cite{liu2016ssd}                       & Det.         & ResNet     & -            & 0.406            \\
MaskRCNN\cite{he2017mask}                  & Seg. \& Det. & R50        & 0.452        & 0.376            \\
AP-MTL\cite{islam2020ap}                   & Seg. \& Det. & Trfmr      & 0.947        & 0.392            \\
MT3DNet(ours)                              & Seg. \& Det. & Trfmr      & 0.966        & 0.451                \\ \bottomrule
\end{tabular}
\end{table}

\section{Experimental Results}
\label{sec:results}
In our experiment, we used the  \textit{Adan} optimizer \cite{xie2023adanadaptivenesterovmomentum} with an initial learning rate of $0.001$ and a weighted decay of $0.02$. We also incorporated a Reduce on Plateau scheduler and an exponential learning rate scheduler to enhance learning dynamics, ensuring efficient convergence and improved performance throughout training, contributing to robustness and effectiveness. Leveraging the publicly available EndoVis18 dataset, comprising 14 videos of porcine surgeries with The Da Vinci Xi \cite{The_da_Vinci_Xi_2017} surgical system, provided meticulous annotations for the entire surgical scene and individual instrument parts. This comprehensive dataset provides the necessary ground truth for segmentation and detection annotations. However, it was missing depth annotations, which are essential for learning depth data. To address this, depth maps were created using the state-of-the-art STL depth estimation model, Depth Anything\cite{depth_anything_v1}, which takes in visual input to infer pixel-wise depth, thus creating high-quality synthetic depth annotations optimized for multi-task learning.

\begin{figure}[h]
    \centering
    \begin{subfigure}{0.24\textwidth}
        \centering
         Input Image
    \end{subfigure}
    \begin{subfigure}{0.24\textwidth}
        \centering
         Depth Map
    \end{subfigure}
    \begin{subfigure}{0.24\textwidth}
        \centering
        3D Reconstruciton
    \end{subfigure}
    \begin{subfigure}{0.24\textwidth}
        \centering
        Segm. \& Det.
    \end{subfigure}
    \par\bigskip
    \begin{subfigure}{0.24\textwidth}
        \centering
        \includegraphics[width=\textwidth]{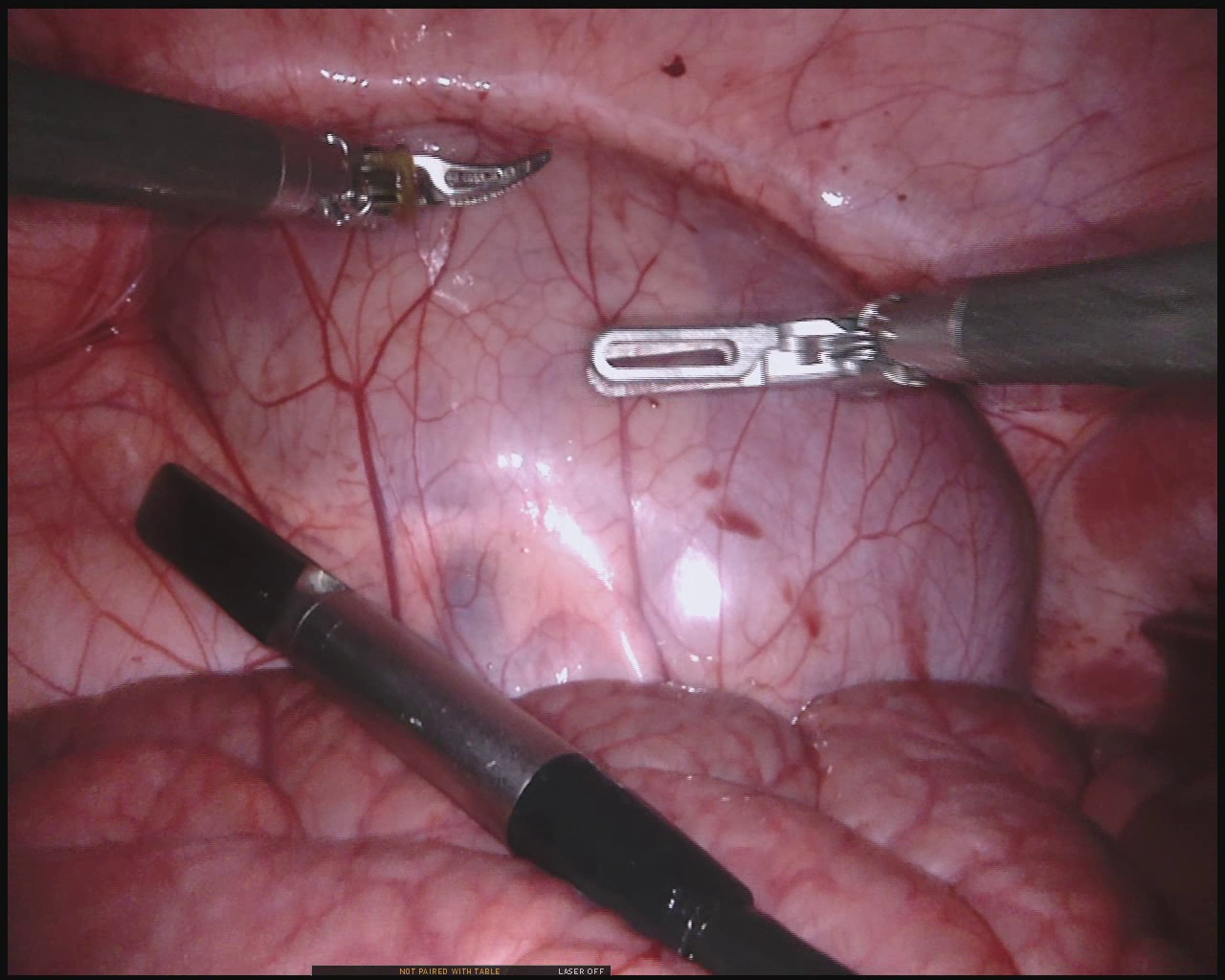}
        \caption{}
    \end{subfigure}
    \begin{subfigure}{0.24\textwidth}
        \centering
        \includegraphics[width=\textwidth]{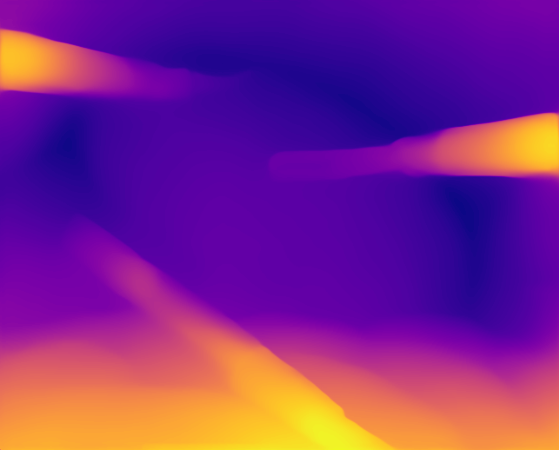}
        \caption{}
    \end{subfigure}
    \begin{subfigure}{0.24\textwidth}
        \centering
        \includegraphics[width=\textwidth]{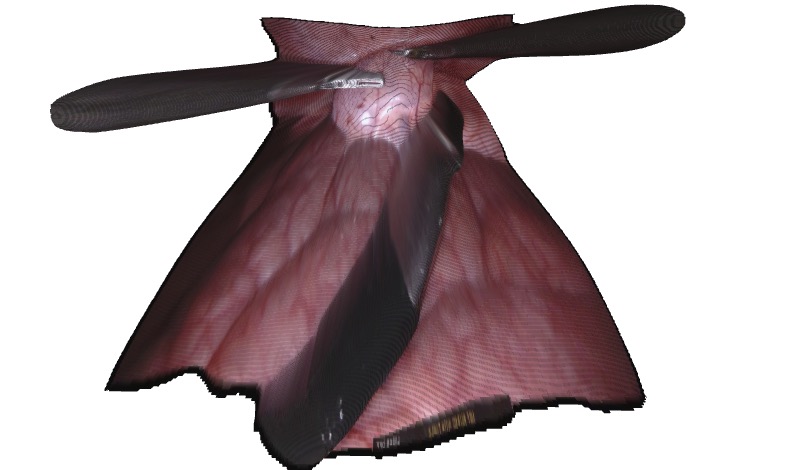}
        \caption{}
    \end{subfigure}
    \begin{subfigure}{0.24\textwidth}
        \centering
        \includegraphics[width=\textwidth]{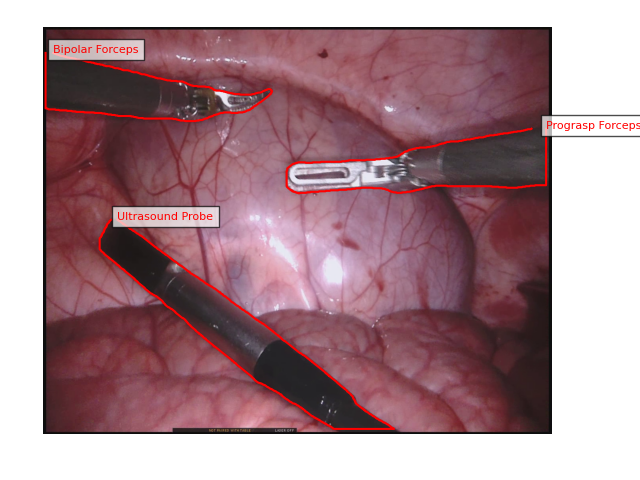}
        \caption{}
    \end{subfigure}
    \par\bigskip
    \begin{subfigure}{0.24\textwidth}
        \centering
        \includegraphics[width=\textwidth]{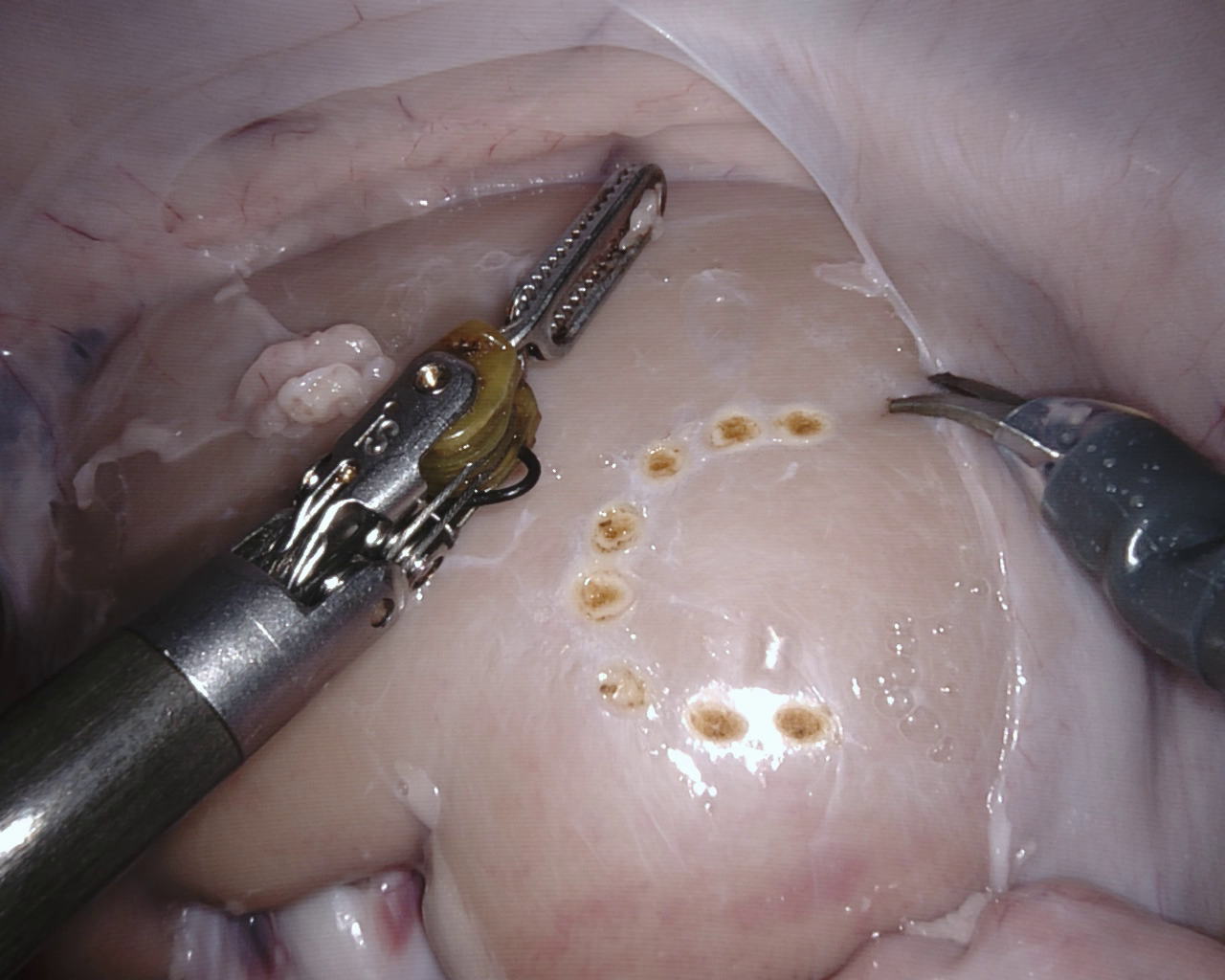}
        \caption{}
    \end{subfigure}
    \begin{subfigure}{0.24\textwidth}
        \centering
        \includegraphics[width=\textwidth]{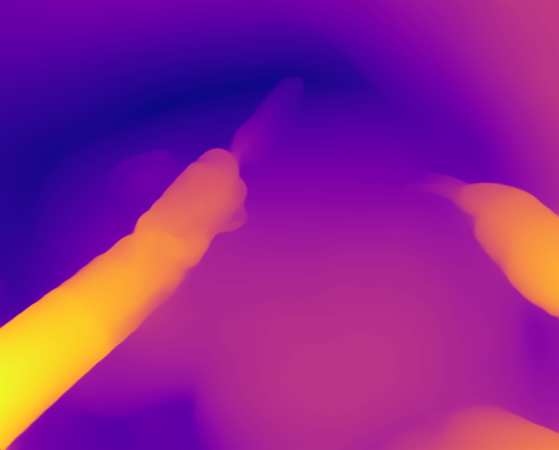}
        \caption{}
    \end{subfigure}
    \begin{subfigure}{0.24\textwidth}
        \centering
        \includegraphics[width=\textwidth]{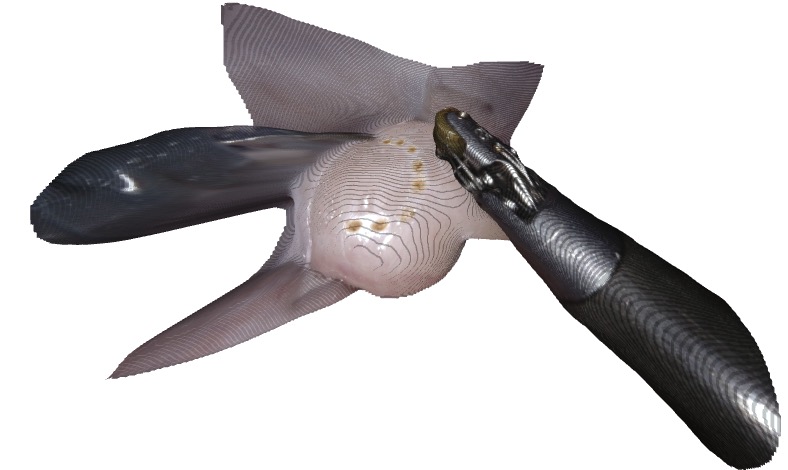}
        \caption{}
    \end{subfigure}
    \begin{subfigure}{0.24\textwidth}
        \centering
        \includegraphics[width=\textwidth]{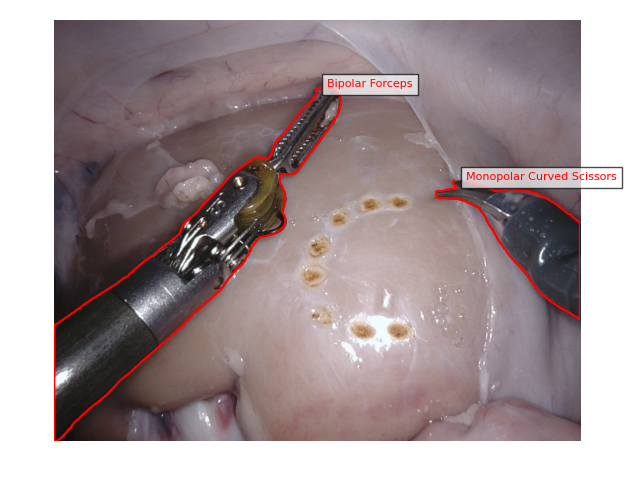}
        \caption{}
    \end{subfigure}
    \par\bigskip
    \begin{subfigure}{0.24\textwidth}
        \centering
        \includegraphics[width=\textwidth]{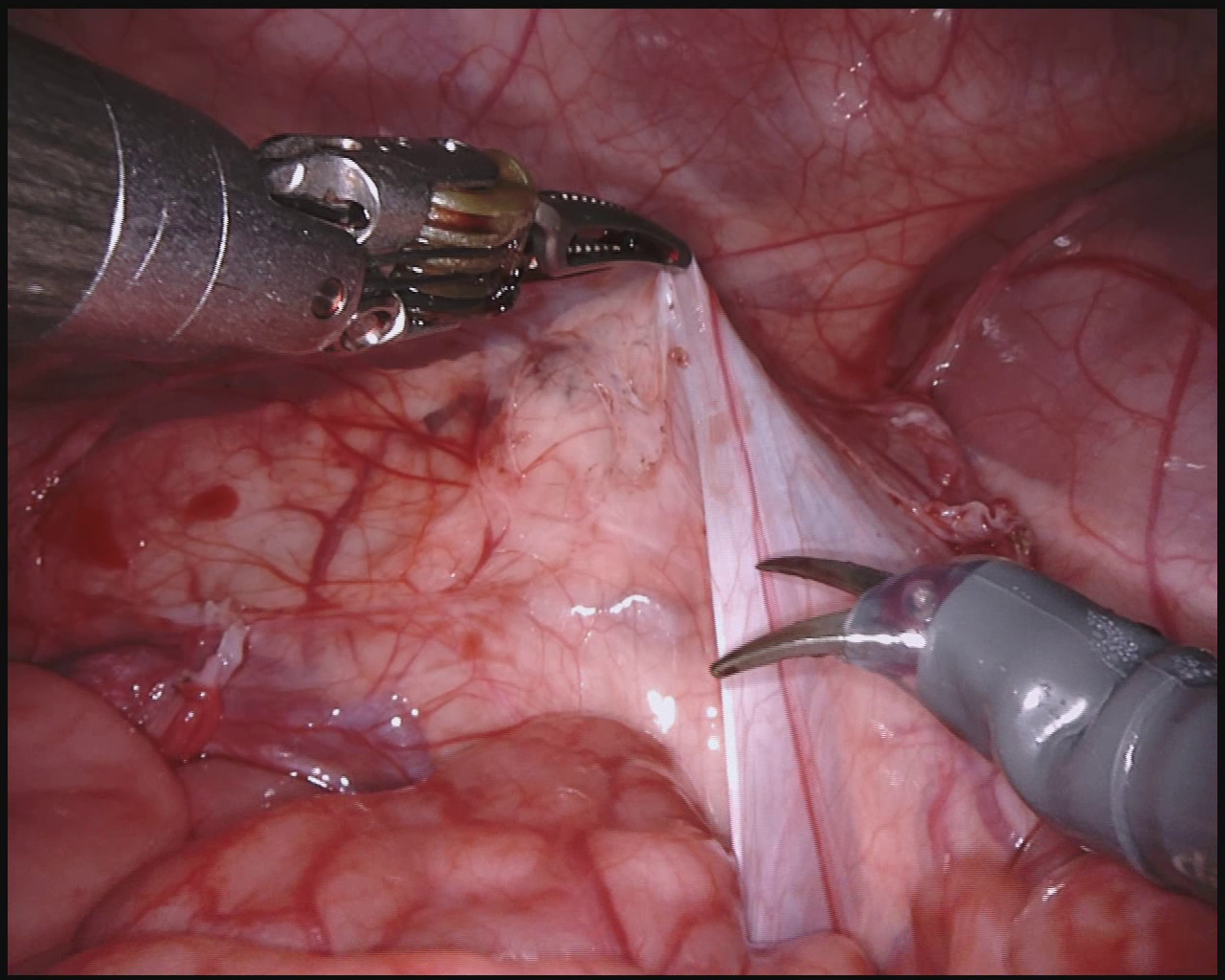}
        \caption{}
    \end{subfigure}
    \begin{subfigure}{0.24\textwidth}
        \centering
        \includegraphics[width=\textwidth]{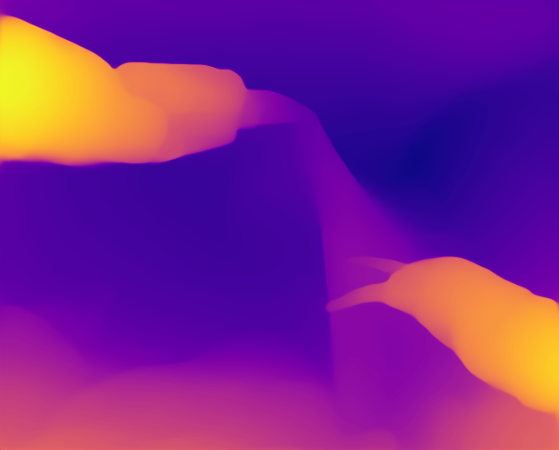}
        \caption{}
    \end{subfigure}
    \begin{subfigure}{0.24\textwidth}
        \centering
        \includegraphics[width=\textwidth]{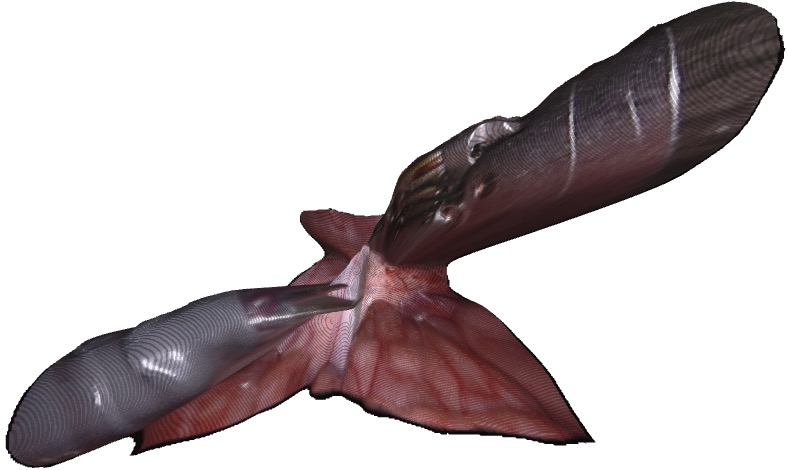}
        \caption{}
    \end{subfigure}
    \begin{subfigure}{0.24\textwidth}
        \centering
        \includegraphics[width=\textwidth]{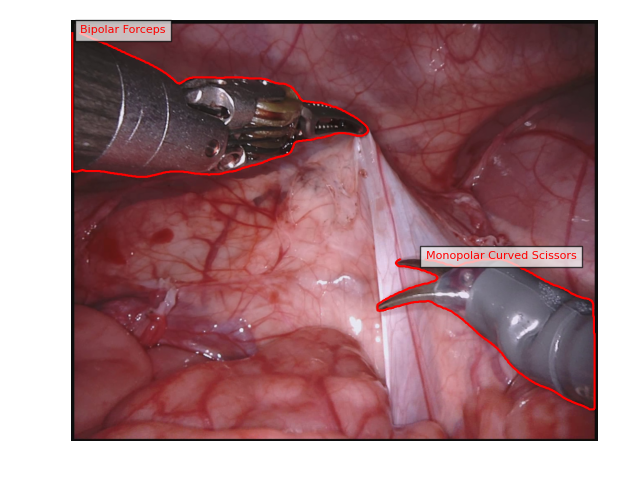}
        \caption{}
    \end{subfigure}
    
\caption{Comprehensive Visualization: RGB Inputs (a, e, i), Depth Estimation (b, f, j) , 3D Representations (c, g, k), and Segmentation \& Detection Outputs (d, h, l)}
\label{fig:metrics}
\end{figure}

In the Table \ref{tab:seg-obj}, performance metrics such Dice, and mAP across different models like TernausNet-11, ERFNet, LinkNet, DeepLabv3+, Yolov3, SSD, MaskRCNN, Ap-MTL and MT3DNet are compared for segmentation (Seg.) and detection (Det.) tasks. The monocular depth estimation prediction gives $2.2$mm MAE; the predicted depth map is later used for 3D reconstruction of the scene. The performance of various models on segmentation and detection tasks is illustrated in Fig. \ref{fig:metrics}.

The ablation study evaluates the impact of incorporating the Adversarial Weight Update algorithm into a Multi-Task Learning (MTL) framework, specifically for image-assisted minimally invasive surgeries. In this framework, the model simultaneously performs segmentation and object detection tasks. The goal is to examine whether introducing optimal weight updates improves performance across these tasks.
\begin{table}[h]
\centering
\caption{Ablation study of our model with and without Adversarial Weight Update algorithm for segmentation (Seg.) and object detection (Det.) tasks, with metrics including Dice, and mAP.}
\label{tab:abl_tab}
\begin{tabular}{@{}ccccc@{}}
\toprule
Model         & Regime       & Arch. & Segmentation & Object detection \\ \midrule
              &              &       & Dice         & mAP              \\ \midrule \midrule
MT3DNet(w/o)  & Seg. \& Det. & Trfmr & 0.921        & 0.377            \\  \midrule
MT3DNet(with) & Seg. \& Det. & Trfmr & 0.966        & 0.451            \\ \bottomrule
\end{tabular}
\end{table}
The study compares the performance of the MTL model with and without the Adversarial Weight Update algorithm. As shown in Table \ref{tab:abl_tab}, the model with the Adversarial Weight Update demonstrates significant improvements over the baseline. Specifically, the Dice coefficient for segmentation has increased by 4.5\%, and the mean Average Precision (mAP) for object detection has improved by 7.4\%. These improvements highlight the effectiveness of the proposed approach in achieving better task-specific performance and overall robustness in multi-task learning.

Future research might be focused on the exploration of the generalizability of MT3DNet to other medical imaging modalities and surgical procedures. Further, exploring real-time implementation and integration into existing surgical systems could open the doors for practical deployment in clinical settings. Another possible area of further research is exploring the potential integration of additional sensory data, such as tactile or force feedback, which would enrich the understanding and guidance capabilities of the system during surgical procedures.
\section{Conclusion}
In this study, a novel multi-task learning framework called MT3DNet was proposed for segmenting, localizing, and estimating depth in image-assisted Minimally Invasive Surgery environments. With the use of Adversarial Weight Update, challenges of optimization inherent with the management of multiple tasks simultaneously were addressed to bring about optimality. Extensive experiments on the EndoVis2018 dataset demonstrated the model's ability to handle all three tasks, which demonstrated the effectiveness of our methodologies. Importantly, the approach enables 3D reconstruction by integrating segmentation, depth estimation, and object detection, which improves surgical scene understanding significantly over prior studies that lack 3D capabilities. The novel approach with the prototype shows potential for use in diverse scenarios with segmentation, detection, and depth annotation. Further research should be done in advancing 3D reconstruction across various different avenues using this approach.

\bibliographystyle{splncs04}
\bibliography{mt3dnet}

\begin{thebibliography}{10}
\providecommand{\url}[1]{\texttt{#1}}
\providecommand{\urlprefix}{URL }
\providecommand{\doi}[1]{https://doi.org/#1}

\bibitem{baby2023forks}
Baby, B., Thapar, D., Chasmai, M., Banerjee, T., Dargan, K., Suri, A., Banerjee, S., Arora, C.: From forks to forceps: A new framework for instance segmentation of surgical instruments. In: Proceedings of the IEEE/CVF Winter Conference on Applications of Computer Vision. pp. 6191--6201 (2023)

\bibitem{baheti2020semantic}
Baheti, B., Innani, S., Gajre, S., Talbar, S.: Semantic scene segmentation in unstructured environment with modified deeplabv3+. Pattern Recognition Letters  \textbf{138},  223--229 (2020)

\bibitem{chaurasia2017linknet}
Chaurasia, A., Culurciello, E.: Linknet: Exploiting encoder representations for efficient semantic segmentation. In: 2017 IEEE visual communications and image processing (VCIP). pp.~1--4. IEEE (2017)

\bibitem{cheng2021deep}
Cheng, T., Li, W., Ng, W.Y., Huang, Y., Li, J., Ng, C.S.H., Chiu, P.W.Y., Li, Z.: Deep learning assisted robotic magnetic anchored and guided endoscope for real-time instrument tracking. IEEE Robotics and Automation Letters  \textbf{6}(2),  3979--3986 (2021)

\bibitem{eddie2020serv}
Eddie''Edwards, P., Psychogyios, D., Speidel, S., Maier-Hein, L., Stoyanov, D.: Serv-ct: A disparity dataset from ct for validation of endoscopic 3d reconstruction. arXiv e-prints pp. arXiv--2012 (2020)

\bibitem{eigen2015predicting}
Eigen, D., Fergus, R.: Predicting depth, surface normals and semantic labels with a common multi-scale convolutional architecture. In: Proceedings of the IEEE international conference on computer vision. pp. 2650--2658 (2015)

\bibitem{he2017mask}
He, K., Gkioxari, G., Doll{\'a}r, P., Girshick, R.: Mask r-cnn. In: Proceedings of the IEEE international conference on computer vision. pp. 2961--2969 (2017)

\bibitem{iglovikov2021ternausnet}
Iglovikov, V.I., Shvets, A.A.: Ternausnet. Computer-aided analysis of gastrointestinal videos pp. 127--132 (2021)

\bibitem{islam2020ap}
Islam, M., Vibashan, V., Ren, H.: Ap-mtl: Attention pruned multi-task learning model for real-time instrument detection and segmentation in robot-assisted surgery. In: 2020 IEEE international conference on robotics and automation (ICRA). pp. 8433--8439. IEEE (2020)

\bibitem{The_da_Vinci_Xi_2017}
James Chi-Yong~Ngu, C.B.S.T., Koh, D.C.S.: The da vinci xi: a review of its capabilities, versatility, and potential role in robotic colorectal surgery. Robotic Surgery: Research and Reviews  \textbf{4},  77--85 (2017). \doi{10.2147/RSRR.S119317}, \url{https://www.tandfonline.com/doi/abs/10.2147/RSRR.S119317}

\bibitem{jiang2019sense}
Jiang, H., Sun, D., Jampani, V., Lv, Z., Learned-Miller, E., Kautz, J.: Sense: A shared encoder network for scene-flow estimation. In: Proceedings of the IEEE/CVF International Conference on Computer Vision. pp. 3195--3204 (2019)

\bibitem{kokkinos2017ubernet}
Kokkinos, I.: Ubernet: Training a universal convolutional neural network for low-, mid-, and high-level vision using diverse datasets and limited memory. In: Proceedings of the IEEE conference on computer vision and pattern recognition. pp. 6129--6138 (2017)

\bibitem{kurmann2021mask}
Kurmann, T., M{\'a}rquez-Neila, P., Allan, M., Wolf, S., Sznitman, R.: Mask then classify: Multi-instance segmentation for surgical instruments. International journal of computer assisted radiology and surgery  \textbf{16}(7),  1227--1236 (2021)

\bibitem{kurmann2017simultaneous}
Kurmann, T., Marquez~Neila, P., Du, X., Fua, P., Stoyanov, D., Wolf, S., Sznitman, R.: Simultaneous recognition and pose estimation of instruments in minimally invasive surgery. In: Medical Image Computing and Computer-Assisted Intervention- MICCAI 2017: 20th International Conference, Quebec City, QC, Canada, September 11-13, 2017, Proceedings, Part II 20. pp. 505--513. Springer (2017)

\bibitem{liu2016ssd}
Liu, W., Anguelov, D., Erhan, D., Szegedy, C., Reed, S., Fu, C.Y., Berg, A.C.: Ssd: Single shot multibox detector. In: Computer Vision--ECCV 2016: 14th European Conference, Amsterdam, The Netherlands, October 11--14, 2016, Proceedings, Part I 14. pp. 21--37. Springer (2016)

\bibitem{pakhomov2019deep}
Pakhomov, D., Premachandran, V., Allan, M., Azizian, M., Navab, N.: Deep residual learning for instrument segmentation in robotic surgery. In: Machine Learning in Medical Imaging: 10th International Workshop, MLMI 2019, Held in Conjunction with MICCAI 2019, Shenzhen, China, October 13, 2019, Proceedings 10. pp. 566--573. Springer (2019)

\bibitem{psychogyios2022msdesis}
Psychogyios, D., Mazomenos, E., Vasconcelos, F., Stoyanov, D.: Msdesis: Multitask stereo disparity estimation and surgical instrument segmentation. IEEE transactions on medical imaging  \textbf{41}(11),  3218--3230 (2022)

\bibitem{redmon2018yolov3}
Redmon, J., Farhadi, A.: Yolov3: An incremental improvement. arXiv preprint arXiv:1804.02767  (2018)

\bibitem{romera2017erfnet}
Romera, E., Alvarez, J.M., Bergasa, L.M., Arroyo, R.: Erfnet: Efficient residual factorized convnet for real-time semantic segmentation. IEEE Transactions on Intelligent Transportation Systems  \textbf{19}(1),  263--272 (2017)

\bibitem{sarikaya2017detection}
Sarikaya, D., Corso, J.J., Guru, K.A.: Detection and localization of robotic tools in robot-assisted surgery videos using deep neural networks for region proposal and detection. IEEE transactions on medical imaging  \textbf{36}(7),  1542--1549 (2017)

\bibitem{stoyanov2010real}
Stoyanov, D., Scarzanella, M.V., Pratt, P., Yang, G.Z.: Real-time stereo reconstruction in robotically assisted minimally invasive surgery. In: Medical Image Computing and Computer-Assisted Intervention--MICCAI 2010: 13th International Conference, Beijing, China, September 20-24, 2010, Proceedings, Part I 13. pp. 275--282. Springer (2010)

\bibitem{wang2021efficient}
Wang, J., Jin, Y., Wang, L., Cai, S., Heng, P.A., Qin, J.: Efficient global-local memory for real-time instrument segmentation of robotic surgical video. In: Medical Image Computing and Computer Assisted Intervention--MICCAI 2021: 24th International Conference, Strasbourg, France, September 27--October 1, 2021, Proceedings, Part IV 24. pp. 341--351. Springer (2021)

\bibitem{xie2023adanadaptivenesterovmomentum}
Xie, X., Zhou, P., Li, H., Lin, Z., Yan, S.: Adan: Adaptive nesterov momentum algorithm for faster optimizing deep models (2023), \url{https://arxiv.org/abs/2208.06677}

\bibitem{depth_anything_v1}
Yang, L., Kang, B., Huang, Z., Xu, X., Feng, J., Zhao, H.: Depth anything: Unleashing the power of large-scale unlabeled data. In: CVPR (2024)

\bibitem{zhao2022trasetr}
Zhao, Z., Jin, Y., Heng, P.A.: Trasetr: track-to-segment transformer with contrastive query for instance-level instrument segmentation in robotic surgery. In: 2022 International Conference on Robotics and Automation (ICRA). pp. 11186--11193. IEEE (2022)

\bibitem{zhou2019real}
Zhou, H., Jagadeesan, J.: Real-time dense reconstruction of tissue surface from stereo optical video. IEEE transactions on medical imaging  \textbf{39}(2),  400--412 (2019)

\end{thebibliography}

\end{document}